\documentclass[10pt, conference]{IEEEtran}
\usepackage[utf8]{inputenc} 
\usepackage{graphicx,url}
\usepackage{multirow}
\usepackage{hhline}
\graphicspath{ {./images/} }
\usepackage{graphicx}
\usepackage{subfigure,amssymb,amsmath}
\usepackage{color,soul}
\usepackage[dvipsnames]{xcolor}
\usepackage{bbm}
\usepackage{tabularx}
\usepackage{float}
\usepackage{booktabs}       
\usepackage{makecell}

\makeatletter
\newif\if@restonecol
\makeatother

\usepackage[linesnumbered,ruled,vlined]{algorithm2e}
\usepackage{algpseudocode}
\usepackage{amsmath}
\usepackage{fancyhdr} 
\usepackage{tabularx}
\usepackage{comment} 

\renewcommand\thetable{\arabic{table}}
\usepackage{tikz,pgfplots}
\usepackage{pgfplots}
\usetikzlibrary{shapes.geometric,backgrounds,patterns, trees}
\usetikzlibrary{3d,decorations.text,shapes.arrows,positioning,fit,backgrounds}
\usetikzlibrary{positioning, decorations.pathmorphing, shapes}
\usetikzlibrary{decorations.pathreplacing}
\usetikzlibrary{shapes.geometric,backgrounds,patterns, trees}
\usetikzlibrary{arrows.meta,
                bending,
                intersections,
                quotes,
                shapes.geometric}
              \usetikzlibrary{automata, positioning}
\usepgfplotslibrary{fillbetween}
\usetikzlibrary{shapes,arrows}
\usetikzlibrary{arrows.meta}
\usetikzlibrary{positioning}
\tikzset{set/.style={draw,circle,inner sep=0pt,align=center}}
\usetikzlibrary{automata, positioning}
  \usetikzlibrary{shapes,shadows}
  \tikzstyle{abstractbox} = [draw=black, fill=white, rectangle,
  inner sep=10pt, style=rounded corners, drop shadow={fill=black,
  opacity=1}]
\tikzstyle{abstracttitle} =[fill=white]
\tikzstyle{block} = [rectangle, draw,
text width=10.5em, text centered, rounded corners, minimum height=4em]
\tikzstyle{line} = [draw, -latex]
\usetikzlibrary{calc,positioning,shapes.geometric}
\usetikzlibrary{arrows.meta,arrows}

\usetikzlibrary{matrix}
\tikzstyle{cblue}=[circle, draw, thin,fill=cyan!20, scale=0.8]
\tikzstyle{qgre}=[rectangle, draw, thin,fill=green!20, scale=0.8]
\tikzstyle{rpath}=[ultra thick, red, opacity=0.4]
\tikzstyle{legend_isps}=[rectangle, rounded corners, thin,
                       fill=gray!20, text=blue, draw]

\tikzstyle{legend_overlay}=[rectangle, rounded corners, thin,
                           top color= white,bottom color=green!25,
                           minimum width=2.5cm, minimum height=0.8cm,
                           pinegreen]
\tikzstyle{legend_phytop}=[rectangle, rounded corners, thin,
                          top color= white,bottom color=cyan!25,
                          minimum width=2.5cm, minimum height=0.8cm,
                          royalblue]
\tikzstyle{legend_general}=[rectangle, rounded ckith'sers, thin,
                          top color= white,bottom color=lavander!25,
                          minimum width=2.5cm, minimum height=0.8cm,
                          violet]
                          \usetikzlibrary{matrix}

\colorlet{myRed}{red!20}
\tikzset{
  rows/.style 2 args={/utils/temp/.style={row ##1/.append style={nodes={#2}}},
    /utils/temp/.list={#1}},
  columns/.style 2 args={/utils/temp/.style={column ##1/.append style={nodes={#2}}},
    /utils/temp/.list={#1}}}
\usetikzlibrary{backgrounds,calc,shadings,shapes.arrows,shapes.symbols,shadows}
\definecolor{switch}{HTML}{006996}

\makeatletter
\pgfkeys{/pgf/.cd,
  parallelepiped offset x/.initial=2mm,
  parallelepiped offset y/.initial=2mm
}
\pgfdeclareshape{parallelepiped}
{
  \inheritsavedanchors[from=rectangle] 
  \inheritanchorborder[from=rectangle]
  \inheritanchor[from=rectangle]{north}
  \inheritanchor[from=rectangle]{north west}
  \inheritanchor[from=rectangle]{north east}
  \inheritanchor[from=rectangle]{center}
  \inheritanchor[from=rectangle]{west}
  \inheritanchor[from=rectangle]{east}
  \inheritanchor[from=rectangle]{mid}
  \inheritanchor[from=rectangle]{mid west}
  \inheritanchor[from=rectangle]{mid east}
  \inheritanchor[from=rectangle]{base}
  \inheritanchor[from=rectangle]{base west}
  \inheritanchor[from=rectangle]{base east}
  \inheritanchor[from=rectangle]{south}
  \inheritanchor[from=rectangle]{south west}
  \inheritanchor[from=rectangle]{south east}
  \backgroundpath{
    \southwest \pgf@xa=\pgf@x \pgf@ya=\pgf@y
    \northeast \pgf@xb=\pgf@x \pgf@yb=\pgf@y
    \pgfmathsetlength\pgfutil@tempdima{\pgfkeysvalueof{/pgf/parallelepiped
      offset x}}
    \pgfmathsetlength\pgfutil@tempdimb{\pgfkeysvalueof{/pgf/parallelepiped
      offset y}}
    \def\ppd@offset{\pgfpoint{\pgfutil@tempdima}{\pgfutil@tempdimb}}
    \pgfpathmoveto{\pgfqpoint{\pgf@xa}{\pgf@ya}}
    \pgfpathlineto{\pgfqpoint{\pgf@xb}{\pgf@ya}}
    \pgfpathlineto{\pgfqpoint{\pgf@xb}{\pgf@yb}}
    \pgfpathlineto{\pgfqpoint{\pgf@xa}{\pgf@yb}}
    \pgfpathclose
    \pgfpathmoveto{\pgfqpoint{\pgf@xb}{\pgf@ya}}
    \pgfpathlineto{\pgfpointadd{\pgfpoint{\pgf@xb}{\pgf@ya}}{\ppd@offset}}
    \pgfpathlineto{\pgfpointadd{\pgfpoint{\pgf@xb}{\pgf@yb}}{\ppd@offset}}
    \pgfpathlineto{\pgfpointadd{\pgfpoint{\pgf@xa}{\pgf@yb}}{\ppd@offset}}
    \pgfpathlineto{\pgfqpoint{\pgf@xa}{\pgf@yb}}
    \pgfpathmoveto{\pgfqpoint{\pgf@xb}{\pgf@yb}}
    \pgfpathlineto{\pgfpointadd{\pgfpoint{\pgf@xb}{\pgf@yb}}{\ppd@offset}}
  }
}

\makeatletter
\tikzset{anchor/.append code=\let\tikz@auto@anchor\relax,
  add font/.code=%
    \expandafter\def\expandafter\tikz@textfont\expandafter{\tikz@textfont#1},
  left delimiter/.style 2 args={append after command={\tikz@delimiter{south east}
    {south west}{every delimiter,every left delimiter,#2}{south}{north}{#1}{.}{\pgf@y}}}}
\tikzstyle{sms} = [rectangle callout, draw, very thick, rounded corners, minimum height=20pt]
\makeatletter
\tikzset{anchor/.append code=\let\tikz@auto@anchor\relax,
  add font/.code=%
    \expandafter\def\expandafter\tikz@textfont\expandafter{\tikz@textfont#1},
  left delimiter/.style 2 args={append after command={\tikz@delimiter{south east}
    {south west}{every delimiter,every left delimiter,#2}{south}{north}{#1}{.}{\pgf@y}}}}
\tikzstyle{sms} = [rectangle callout, draw,very thick, rounded corners, minimum height=20pt]
\usetikzlibrary{positioning,calc}

\tikzset{
  mybackground9/.style={execute at end picture={
        \begin{scope}[on background layer]
          \draw[black,fill=Black!5!Sepia!1,rounded corners=6ex] (current bounding box.south west)
                    rectangle (current bounding box.north east);
          \node[draw,fill=white,ellipse,anchor=west,inner sep=1pt,minimum width=4ex] at (current bounding box.north
                   west){#1};
        \end{scope}
    }},
}

\tikzset{
  mybackground10/.style={execute at end picture={
        \begin{scope}[on background layer]
          \draw[black] (current bounding box.south west)
                    rectangle (current bounding box.north east);
          \node[draw,fill=white,ellipse,anchor=west,inner sep=1pt,minimum width=4ex] at (current bounding box.north
                   west){#1};
        \end{scope}
    }},
}

\tikzset{l3 switch/.style={
    parallelepiped,fill=switch, draw=white,
    minimum width=0.75cm,
    minimum height=0.75cm,
    parallelepiped offset x=1.75mm,
    parallelepiped offset y=1.25mm,
    path picture={
      \node[fill=white,
        circle,
        minimum size=6pt,
        inner sep=0pt,
        append after command={
          \pgfextra{
            \foreach \angle in {0,45,...,360}
            \draw[-latex,fill=white] (\tikzlastnode.\angle)--++(\angle:2.25mm);
          }
        }
      ]
       at ([xshift=-0.75mm,yshift=-0.5mm]path picture bounding box.center){};
    }
  },
  ports/.style={
    line width=0.3pt,
    top color=gray!20,
    bottom color=gray!80
  },
  rack switch/.style={
    parallelepiped,fill=white, draw,
    minimum width=1.25cm,
    minimum height=0.25cm,
    parallelepiped offset x=2mm,
    parallelepiped offset y=1.25mm,
    xscale=-1,
    path picture={
      \draw[top color=gray!5,bottom color=gray!40]
      (path picture bounding box.south west) rectangle
      (path picture bounding box.north east);
      \coordinate (A-west) at ([xshift=-0.2cm]path picture bounding box.west);
      \coordinate (A-center) at ($(path picture bounding box.center)!0!(path
        picture bounding box.south)$);
      \foreach \x in {0.275,0.525,0.775}{
        \draw[ports]([yshift=-0.05cm]$(A-west)!\x!(A-center)$)
          rectangle +(0.1,0.05);
        \draw[ports]([yshift=-0.125cm]$(A-west)!\x!(A-center)$)
          rectangle +(0.1,0.05);
       }
      \coordinate (A-east) at (path picture bounding box.east);
      \foreach \x in {0.085,0.21,0.335,0.455,0.635,0.755,0.875,1}{
        \draw[ports]([yshift=-0.1125cm]$(A-east)!\x!(A-center)$)
          rectangle +(0.05,0.1);
      }
    }
  },
  server/.style={
    parallelepiped,
    fill=white, draw,
    minimum width=0.35cm,
    minimum height=0.75cm,
    parallelepiped offset x=3mm,
    parallelepiped offset y=2mm,
    xscale=-1,
    path picture={
      \draw[top color=gray!5,bottom color=gray!40]
      (path picture bounding box.south west) rectangle
      (path picture bounding box.north east);
      \coordinate (A-center) at ($(path picture bounding box.center)!0!(path
        picture bounding box.south)$);
      \coordinate (A-west) at ([xshift=-0.575cm]path picture bounding box.west);
      \draw[ports]([yshift=0.1cm]$(A-west)!0!(A-center)$)
        rectangle +(0.2,0.065);
      \draw[ports]([yshift=0.01cm]$(A-west)!0.085!(A-center)$)
        rectangle +(0.15,0.05);
      \fill[black]([yshift=-0.35cm]$(A-west)!-0.1!(A-center)$)
        rectangle +(0.235,0.0175);
      \fill[black]([yshift=-0.385cm]$(A-west)!-0.1!(A-center)$)
        rectangle +(0.235,0.0175);
      \fill[black]([yshift=-0.42cm]$(A-west)!-0.1!(A-center)$)
        rectangle +(0.235,0.0175);
    }
  },
}
\pgfplotsset{compat=1.16}
\usetikzlibrary{calc, shadings, shadows, shapes.arrows}

\tikzset{%
  interface/.style={draw, rectangle, rounded corners, font=\LARGE\sffamily},
  ethernet/.style={interface, fill=yellow!50},
  serial/.style={interface, fill=green!70},
  speed/.style={sloped, anchor=south, font=\large\sffamily},
  route/.style={draw, shape=single arrow, single arrow head extend=4mm,
    minimum height=1.7cm, minimum width=3mm, white, fill=switch!20,
    drop shadow={opacity=.8, fill=switch}, font=\tiny}
}
\makeatletter
\pgfdeclareradialshading[tikz@ball]{cloud}{\pgfpoint{-0.275cm}{0.4cm}}{%
  color(0cm)=(tikz@ball!75!white);
  color(0.1cm)=(tikz@ball!85!white);
  color(0.2cm)=(tikz@ball!95!white);
  color(0.7cm)=(tikz@ball!89!black);
  color(1cm)=(tikz@ball!75!black)
}
\tikzoption{cloud color}{\pgfutil@colorlet{tikz@ball}{#1}%
  \def\tikz@shading{cloud}\tikz@addmode{\tikz@mode@shadetrue}}
\makeatother

\tikzset{my cloud/.style={
     cloud, draw, aspect=2,
     cloud color={gray!5!white}
  }
}

\usepackage{array}
\newcommand{\thickhline}{%
    \noalign {\ifnum 0=`}\fi \hrule height 1pt
    \futurelet \reserved@a \@xhline
}
\newcolumntype{"}{@{\hskip\tabcolsep\vrule width 1pt\hskip\tabcolsep}}

\IEEEoverridecommandlockouts
\IEEEpubid{\makebox[\columnwidth]{978-3-903176-31-7~\copyright~2020 IFIP\hfill} \hspace{\columnsep}\makebox[\columnwidth]{ }}

\usepackage{etoolbox}
\makeatletter
\patchcmd{\@makecaption}
  {\scshape}
  {}
  {}
  {}
\makeatother

\begin{document}


\title{A Modular Framework for Rapidly Building Intrusion Predictors}

\author{\IEEEauthorblockN{Xiaoxuan Wang \IEEEauthorrefmark{2} and
 Rolf Stadler\IEEEauthorrefmark{2}}

 \IEEEauthorblockA{\IEEEauthorrefmark{2}
Dept. of Computer Science, KTH Royal Institute of Technology, Sweden
 }
  \newline
Email: \{xiaoxuan, stadler\}@kth.se
\\
}

\maketitle

\thispagestyle{plain}
\pagestyle{plain}

\begin{abstract}
\label{sec:abstract} 
We study automated intrusion prediction in an IT system using statistical learning methods. The focus is on developing online attack predictors that detect attacks in real time and identify the current stage of the attack. While such predictors have been proposed in the recent literature, these works typically rely on constructing a monolithic predictor tailored to a specific attack type and scenario. Given that hundreds of attack types are cataloged in the MITRE framework, training a separate monolithic predictor for each of them is infeasible. In this paper, we propose a modular framework for rapidly assembling online attack predictors from reusable components. The modular nature of a predictor facilitates controlling key metrics like timeliness and accuracy of prediction, as well as tuning the trade-off between them. Using public datasets for training and evaluation, we provide many examples of modular predictors and show how an effective predictor can be dynamically assembled during training from a network of modular components.

\end{abstract}


\section{Introduction}
\label{sec:introduction}

Traditional intrusion detection systems (IDS), such as Snort \cite{snort3} or Suricata \cite{suricata}, rely on rule-based configurations that are manually crafted and maintained by domain experts. The growing complexity and rapid evolution of IT systems make the maintenance of these rules increasingly challenging and time-consuming. As a response, research efforts into automated cyberdefence have started, based on the idea that attack patterns can be dynamically learned. The rules are no longer defined by humans, but automatically inferred from observing systems under attack.

Over the last decade, various approaches have been proposed for automated cyberdefence, most of them based on statistical learning, e.g., \cite{shawly2019architectures,ghafir2019hidden,choubisa2022simple,laghrissi2021intrusion}. We follow this direction in the paper. We are specifically interested in predicting the stage of an ongoing attack in real time, based on current and earlier observations of an IT system. An attack stage can include several distinct actions by an attacker, as explained in \cite{navarro2018}.
We use the term \emph{intrusion prediction} instead of the traditional term \emph{intrusion detection} to stress that we study methods with probabilistic output. Attack stage prediction contributes to situational awareness of a human decision maker or an automated defender agent, enabling them to take appropriate response action. This paper focuses on intrusion prediction, leaving out intrusion response, which we have addressed in other works, e.g., \cite{hammar2022intrusion,hammar2023learning,hammar2024optimal}. 

The MITRE ATT\&CK framework is a public, comprehensive, and regularly updated knowledge base that documents known cyberattacks in enterprise environments \cite{mitre}. It describes the tactics, techniques, and procedures used by cyber adversaries across different stages of an attack, and it categorizes attacker behavior into high-level tactics, like initial access or lateral movement, and detailed techniques, such as phishing or credential dumping. For several attacks included in the MITRE knowledge base, researchers have proposed attack stage predictors using machine learning techniques, e.g., \cite{shawly2019architectures,holgado2017real,liao2024multi,zhou2021detecting}. These works discuss predictors that have a monolithic structure and are trained for a specific attack and specific environmental conditions of the IT system. While such predictors have proven effective for the specific purpose and environment they have been designed for, the overall approach of building attack stage predictors becomes impractical, considering the MITRE knowledge base contains hundreds of attacks.

In this paper, we address this issue and present a framework to rapidly build and evaluate attack stage predictors for online intrusion prediction. A predictor is composed of reusable components, which form a chain (or a directed graph) that processes an event stream. The stream entering the chain contains events that are directly observable in the IT system, and the stream leaving the chain consists of attack stage predictions. The modular nature of a predictor facilitates controlling key metrics like timeliness and accuracy of prediction, as well as tuning the trade-off between them, in order to adapt the predictor to a specific use case. Finally, we discuss the issue of efficiently building an effective modular predictor considering all possible candidate predictors that can be constructed from an increasing set of components.


We make the following contributions with this paper:
\begin{itemize}
    \item We propose a modular framework for rapidly constructing online attack predictors from reusable components. The predictors support online, real-time prediction. 
    \item The framework integrates existing systems like Snort as components, thereby leveraging their strengths for more accurate or more efficient predictions. It is extensible through adding new components, e.g., a component that performs an embedding function or a statistical analysis function.
    \item We demonstrate the validity of the framework and illustrate its properties by providing many examples of modular predictors, which we train and evaluate on public datasets. 
    \item We demonstrate the controllability of the built predictors with respect to key metrics, including prediction accuracy, timeliness, computational overhead, and sample efficiency.
    \item We describe and evaluate an approach to efficiently find an effective predictor from a set of candidate predictors during training.
\end{itemize}

The novelty of this paper lies in the architectural concept of assembling attack stage predictors from reusable components. It addresses the problem of building effective attack stage predictors for the large and increasing number of attacks on IT systems. Also, it facilitates tuning key metrics of a predictor. Our work improves on many solutions proposed in the related literature, which generally focus on constructing a single, monolithic predictor for offline use designed to fulfill the requirements and constraints of a given scenario.

\section{Use Case}
\label{sec:use_case}

\begin{figure}[h]
  \centering
    \scalebox{0.22}{      \includegraphics{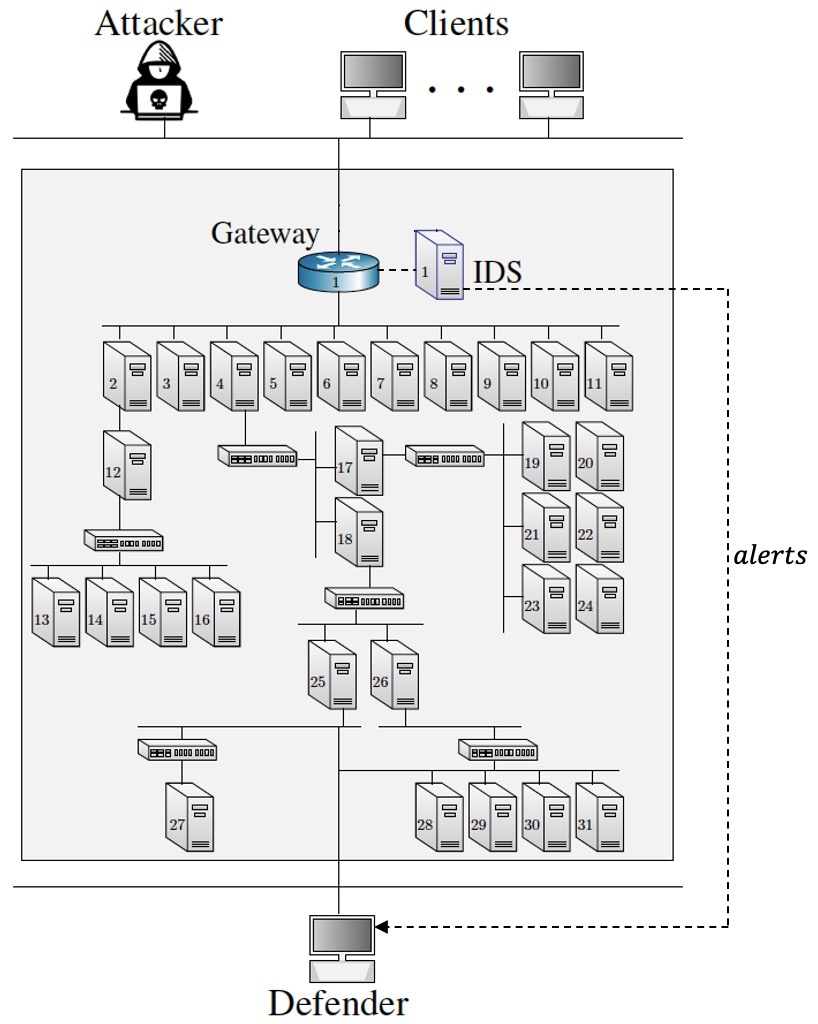}
    }
    \caption{The IT infrastructure in the intrusion prediction use case.}
    \label{fig:IT}
  \end{figure}

We describe a general use case that will help illustrate the framework for intrusion prediction we propose in this paper. We consider the IT infrastructure of an enterprise as shown in Figure \ref{fig:IT}. The infrastructure consists of network computing and storage components, which provide services to clients. Clients access the services through a gateway. This gateway is also accessible to an attacker, whose goal is to infiltrate the infrastructure and perform a multi-stage attack. A defender manages the infrastructure and is responsible for intrusion prediction. The defender receives alerts in real time from an intrusion detection system and from other monitoring functions. Based on this information, the defender predicts whether or not an attack is ongoing and, if so, in which stage the attack is.

\section{Problem formulation and approach}
\label{sec:problem_formulation}

\begin{figure}[h]
  \centering
    \scalebox{0.22}{      
    \includegraphics{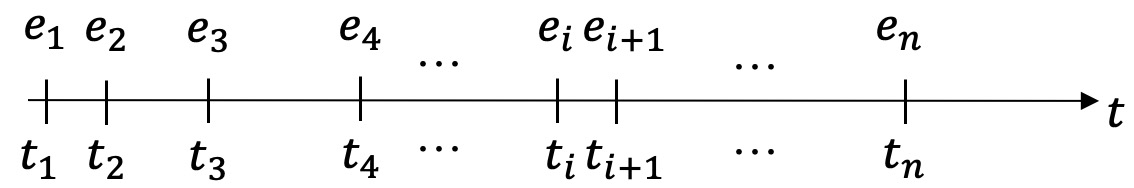}
    }
    \caption{Event stream observed by the defender.}
    \label{fig:stream}
  \end{figure}

We consider an attack from the defender’s point of view. The defender observes a stream of events (e.g., alerts). Each event $e_i$ occurs at a discrete point in time $t_i$, as shown in Figure \ref{fig:stream}. An event has a value, which can be multidimensional, and a timestamp. An online predictor, operated by the defender, processes this stream of events in real time. The predictor iteratively solves a state estimation problem.  It estimates the attack stage based on the observed sequence of events. More precisely, it estimates the attack stage at the current time $t_n$, based on the observations $e_1$, …, $e_n$. Note that the events observed are not necessarily caused by attacker actions. They can be the result of the activity of clients, who access the services of the IT system (see Figure \ref{fig:IT}). Often, most of the events in an event stream are caused by clients and not by the attacker.

\begin{figure*}[t]
    \centering   
    \includegraphics[width=0.9\textwidth]{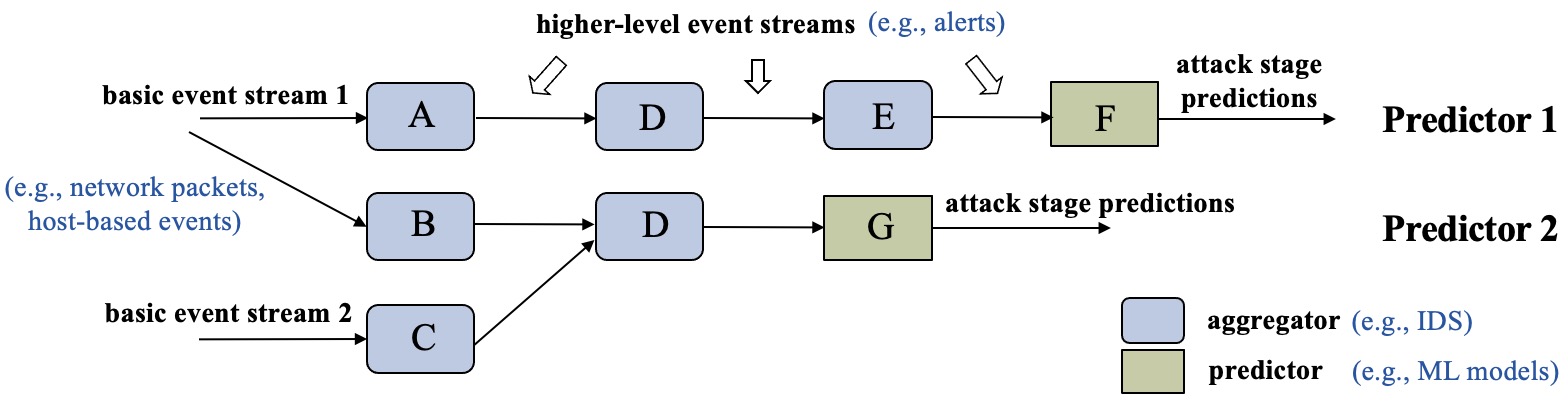} 
    \caption{A generic example of two attack stage predictors built from components. Basic event stream 1 and component D are shared by both predictors.}
    \label{fig:framework}
\end{figure*}

The problem we address in this paper is how to \emph{rapidly build attack stage predictors for a range of attacks} like those described in the MITRE ATT\&CK framework. We advocate the need for rapid construction and training due to the large and increasing number of known attacks on enterprise systems. A predictor should perform real-time prediction. Also, it should be tunable to operate effectively given the requirements and constraints of a specific use case. Building an effective predictor for a scenario requires the capability to control the trade-offs between various metrics. Maintaining the appropriate balance between prediction accuracy, training overhead, sample efficiency, and the number of monitored data sources is essential for achieving both effectiveness and practicality for intrusion prediction. 

Our approach to solving this problem centers around building \emph{a modular framework of reusable components} from which predictors can be quickly assembled. All components operate on event streams; they consume and produce event streams. An existing intrusion detection function or system can be integrated into the framework by encapsulating it into a component. In addition, we build components that implement a variety of methods for event aggregation across time and across value spaces. By adding new components, the framework becomes extensible.  

We control key metrics and tune trade-offs for a modular predictor through various techniques applied in the scenarios of Section \ref{sec:build}. For instance, we control the number and type of input event streams to regulate the monitoring and operational overhead in Scenario 1. We control the timeliness of prediction by setting the window size of an event aggregator component in Scenario 2. We use pretrained components to increase sample efficiency in Scenario 3. Finally, we present a sample algorithm that selects a predictor from a potentially large set of candidate predictors. We control the selection process of the winning predictor by tuning the trade-off between prediction accuracy and training overhead in Scenario 4.

In this paper, we attempt to show the validity of our approach. We illustrate the properties of the framework by providing many examples of modular predictors, which we train and evaluate on public datasets. We demonstrate the controllability of these predictors and show how they can be adapted to the requirements and constraints of specific use cases.

In this paper, we use the terms \emph{event} and \emph{observation} as synonyms. Also, we use the term \emph{alert} for an event produced by a traditional intrusion detection system like Snort.

\section{Our framework for composing online predictors}
\label{sec:framework}

We build predictors from a set of reusable components. Each such component processes one or more incoming event streams in an online fashion, and it produces one or more (identical) outgoing event streams. We think of a component as a node in a directed graph where links that end at that node represent incoming event streams and links that originate at that node represent outgoing event streams. Components are put together to form a directed graph where information flows from the leaves to the root(s) of the graph.

Figure \ref{fig:framework} shows a generic example of two attack stage predictors built from components. Basic event stream 1 and component D are shared by both predictors. Figures~\ref{fig:example1},
\ref{fig:example2}, \ref{fig:example3}, and \ref{fig:example4} give specific examples of online attack stage predictors. 
On the left side of Figure \ref{fig:framework}, we find (basic) event streams formed from events that occur in the system we observe. An event can represent the arrival of a network packet at a network gateway, the establishment of a network flow on a network link, a login attempt on a server, etc. One or more event streams are consumed by an event aggregator, which produces a stream of (higher-level) events that serves as input to the next component on the graph. At the root of the graph is the predictor component, whose output is a sequence of attack stage predictions.

A simple way of composing a modular predictor is to choose a set of components and arrange them in a directed chain. The basic event stream is the input to the first component of the chain, and the attack stage predictions form the output of the last component. In our work to date, we found modular predictors with 3-4 components forming a chain very effective, and most examples in this paper have this structure.

The components shown in Figure \ref{fig:framework} vary significantly in internal complexity. A simple component we found useful implements a sliding window that aggregates $n$ consecutive events into a single event. An example of a more complex component processes $n$ consecutive events using a Long Short-Term Memory (LSTM) encoder. Additionally, we include existing systems like Snort as event aggregators in our framework. Such aggregators exhibit high internal complexity and have been shown to be highly effective as predictor components. 

We envision that each component implements a specific concept or task of general use. We have already built some types of components; others are planned for the near future. An example is the above-mentioned window-based event aggregator, which allows controlling the event rate of the output stream. A different component can perform dimensionality reduction for events with high-dimensional structure, e.g., through feature selection. The inverse task can be realized by a component that embeds events in a high-dimensional space to achieve generalization properties of the predictor, e.g., by using a Graph Neural Network (GNN). Further, a component can implement a clustering technique that maps multi-dimensional events into a finite set of categories, a component can perform time-frequency analysis on an event stream, etc.


In the predictors we built, the rate of events decreases from left to right
while the level of abstraction an event represents increases in Figure \ref{fig:framework}. In fact, the input stream of a left-most component consists of basic events at a high rate, while the output stream of a right-most component is a stream of predictions at a lower rate.

When training a modular predictor, some of its components can be pretrained and thus are kept fixed, while the parameters of other components are tuned. In the examples shown in Section \ref{sec:build}, components at the beginning of a chain are fixed and components towards the end of the chain are tuned during training.  

When we consider a large collection of components to build a modular predictor, we encounter a combinatorial explosion of possible chains for this predictor. Training all possible chains as described above and then selecting the best-performing ones becomes infeasible. We therefore need a heuristic method that identifies an effective predictor with reasonable overhead. In Section \ref{subsec:example4}, we introduce a simple heuristic method that gradually selects a predictor from a shrinking set of candidates during training. A more thorough study of this issue is planned for future work.

\section{Experimental setup}
\label{sec:experiments}

\subsection{Datasets with intrusion attacks}
\label{subsec:data}

We train and evaluate online predictors using three public datasets, which contain multi-stage attacks and are labeled. 

The \textbf{AIT} dataset \cite{landauer2024introducing} includes a multi-stage attack comprising a variety of scanning and exploitation activities, which we call the AIT attack. The attack begins with reconnaissance activities, which include several attack stages, using tools such as Nmap for host and service discovery, Dirb for directory enumeration, and WPScan to analyze an intranet server running WordPress. Reconnaissance is followed by the exploitation of a vulnerable WordPress plugin to upload a webshell, password cracking, the installation of a reverse shell, and ultimately, privilege escalation. Nine attack stages are identified in the dataset, labeled as Normal, Network Scan, Service Scan, WPScan, Dirb Scan, Webshell, Reverse Shell, Password Cracking, and Privilege Escalation. In this paper, `Normal' denotes a special attack stage indicating that the intrusion has not yet begun. The dataset consists of eight scenarios, each containing an attack with the same sequence of nine attack stages, but with varying background traffic and scan intensity. The data is in the form of three alert traces generated from Wazuh \cite{wazuh}, Suricata, and AMiner \cite{aminer}, with timestamps provided at the second level.

The \textbf{MSCAD} dataset \cite{almseidin2022generating} includes a multi-stage attack comprising two distinct scenarios, which we call the MSCAD attack. First, the attacker launches a DDoS attack which progresses through three consecutive stages: port scanning, an application-layer DDoS using the Slowloris technique, and a volume-based DDoS using ICMP flood. Following this, the attacker conducts a password cracking attack that begins with web crawling, and then brute-force password cracking. Six attack stages are identified in this dataset, labeled as Normal, Port Scan, App-based DDoS, Volume-based DDoS, Web Crawling, and Password Cracking. The dataset is in the form of a network packet capture (PCAP) file.

The \textbf{CICIDS} dataset \cite{sharafaldin2018toward} captures five days of network traffic, recorded from 9 a.m. on Monday, July 3, 2017, to 5 p.m. on Friday, July 7, 2017. For our analysis, we focus on the data collected on Thursday, July 6, 2017, which encompasses five stages: Normal, Web Attack-Brute Force, Web Attack-XSS, Web Attack-SQL Injection, and Infiltration. In this paper, we call this attack sequence the CICIDS attack. The dataset is in the form of a PCAP file.

\subsection{Evaluation metrics}

We evaluate the predictors we build using performance metrics for prediction accuracy and overhead. We measure the accuracy of a predictor through the fraction of correctly predicted attack stages (accuracy score) and the macro F1-score. We use the window size of the event aggregator component, i.e., the number of events contained in a window, as a measure for the timeliness of predictions. We evaluate the training overhead by measuring the computing time (on a laptop). Finally, we measure sample efficiency by the amount of data available for training the predictor.

\section{Building and training predictors}
\label{sec:build}
We present four scenarios, for which we build 12 predictors using our framework. In each scenario, we control two metrics, one of which is the prediction accuracy, and we study the trade-off between the two metrics (see Table \ref{tab:scenario_design}). Our component-based framework facilitates the control of performance metrics, as these metrics are often associated with specific components.

\begin{table}[h]
\centering
\caption{Summary of the designed experimental scenarios.}
\resizebox{0.48\textwidth}{!}{
\begin{tabular}{l c c r}
\hline
\textbf{Scenario}&
\textbf{Dataset} & \textbf{Trade-off} & \textbf{Control Parameter} \\
\hline
1 & AIT & monitoring overhead vs. acc & \# incoming event streams  \\
2 & MSCAD & prediction time vs. acc & sliding window size \\
3 & MSCAD & sample efficiency vs. acc & \# pretrained components \\
4 & CICIDS & training overhead vs. acc & \# training folds\\
\hline
\end{tabular}
}
\label{tab:scenario_design}
\end{table}



\subsection{Scenario 1: controlling monitoring overhead vs. accuracy}
\label{subsec:example1}
In this scenario, we build four online predictors for the AIT attack and demonstrate the trade-off between the monitoring setup and overhead on the one side and the prediction accuracy on the other side. We control the monitoring overhead by choosing the number and types of incoming event streams.

\begin{figure}[h]
  \centering
    \scalebox{0.19}{      \includegraphics{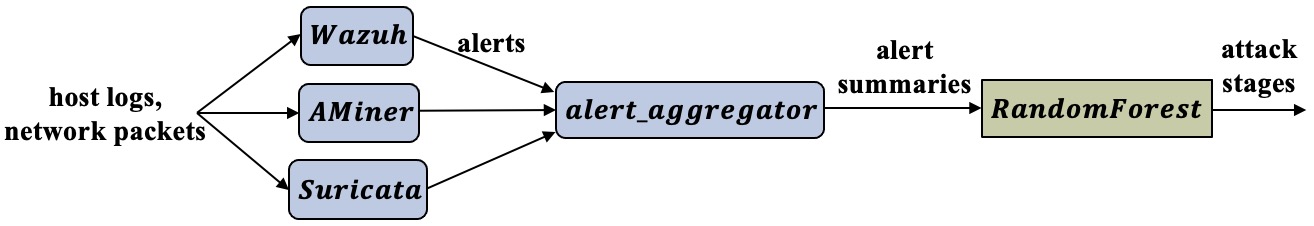}
    }
    \caption{Scenario 1: attack predictors for the AIT attack.}
    \label{fig:example1}
  \end{figure}

The first predictor processes solely the alert stream from Wazuh (W); the second combines the alert streams from Wazuh and Suricata (W+S); the third combines the alert streams from Wazuh and AMiner (W+A); and the fourth integrates the alert streams from all three IDSs, Wazuh, AMiner, and Suricata (W+A+S). Figure \ref{fig:example1} shows the components we use to build the predictors and how they are connected. Streams of network and host events are processed by the three IDSs, Wazuh, Aminer, and Suricata. An alert aggregator component processes one or more alert streams and produces a statistical summary of alerts occurring within a one-second interval. The summary includes the number of alerts from each source and for each alert type. A random forest classifier component processes the stream of alert summaries and outputs a stream of attack stage predictions.

We use four AIT scenario traces for training and the remaining four traces for evaluation. Table \ref{tab:example1} presents the accuracy and F1-score for predicting attack stages of the four predictors. The reported values represent the mean over 10 runs with different random seeds. As discussed, choosing additional data sources (alert streams) improves the prediction accuracy, most notably when expanding from Wazuh alerts to including alerts from Wazuh and AMiner. However, the inclusion of additional data sources comes at the cost of overhead from monitoring and running the IDSs. We note that integrating Suricata alerts does not lead to further improvement. This may be due to Wazuh already processing logs from Suricata, resulting in redundant instead of complementary information \cite{landauer2024introducing}.

\begin{table}[h]
\centering
\caption{Performance of the four AIT attack predictors in Figure \ref{fig:example1}. W stands for Wazuh, S for Suricata, A for AMiner.}
\label{tab:example1}
\begin{tabular}{l c r}
\hline
\textbf{Alert stream used}    & \textbf{Accuracy}    & \textbf{F1-score} \\ \hline
W      & 0.835 & 0.305         \\ 
W+S    & 0.828 & 0.322        \\ 
\textbf{W+A}    & \textbf{0.874} & \textbf{0.530}         \\ 
W+A+S  & 0.870 & 0.526        \\ \hline
\end{tabular}%
\end{table}

\subsection{Scenario 2: controlling prediction time vs. accuracy}
\label{subsec:example2}

In this scenario, we build a modular predictor with seven control settings for the MSCAD attack and demonstrate the trade-off between the timeliness of prediction and the prediction accuracy. By varying the window size of the sliding window component, we regulate the balance between a fast but less accurate predictor and a slower yet more accurate one.

\begin{figure}[h]
  \centering
    \scalebox{0.16}{      \includegraphics{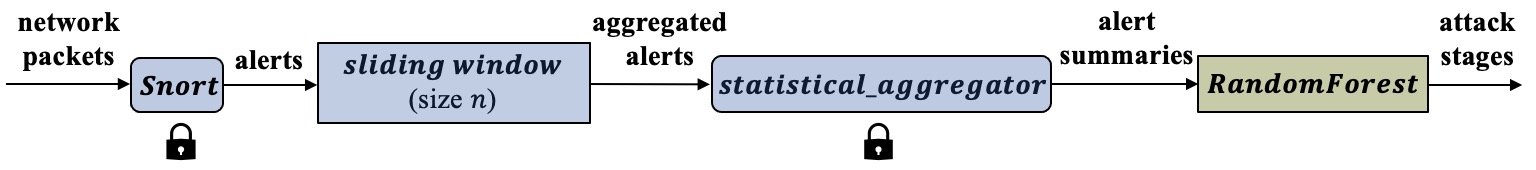}
    }
    \caption{Scenario 2: attack predictor for the MSCAD attack.}
    \label{fig:example2}
  \end{figure}

Figure \ref{fig:example2} shows the components used to build the predictor and how they are connected. The Snort IDS component employs the snortrules-snapshot-31470 ruleset. It processes the stream of network packet events and produces alerts that are consumed by the sliding window component, whose output is sent to a statistical aggregator. The statistical aggregator aggregates alerts within a window, including the number of alerts per type and priority level, as well as the average alert inter-arrival time. We vary the sliding window size: a larger window captures more information but introduces longer prediction times, whereas a smaller window provides less information but enables faster predictions. During training, the Snort component and the statistical aggregator component are kept fixed, while the sliding window size and the random forest parameters are tuned.

\begin{figure}[h]
  \centering
    \scalebox{0.35}{      \includegraphics{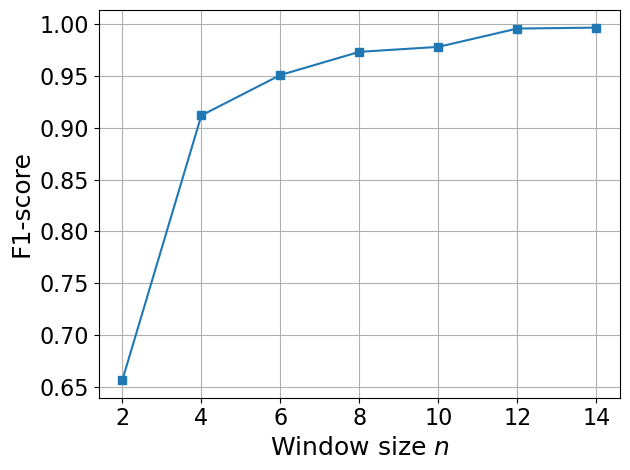}
    }
    \caption{Prediction accuracy of the MSCAD attack predictor for seven different window sizes. A small window size corresponds to a shorter prediction time.}
    \label{fig:result2}
  \end{figure}

Figure \ref{fig:result2} shows that the prediction accuracy improves as the window size increases, particularly when the size expands from 2 to 4. This gain in accuracy comes at the expense of longer prediction times.
  
\subsection{Scenario 3: controlling sample efficiency vs. accuracy}
\label{subsec:example3}

In this scenario, we build three modular predictors for the MSCAD attack with three control settings and demonstrate the trade-off between the sample efficiency and the prediction accuracy. By varying the number of fixed (i.e., pretrained) components of the predictor, we regulate the achievable prediction accuracy in function of the size of the available training data.

\begin{figure}[h]
  \centering
    \scalebox{0.17}{      \includegraphics{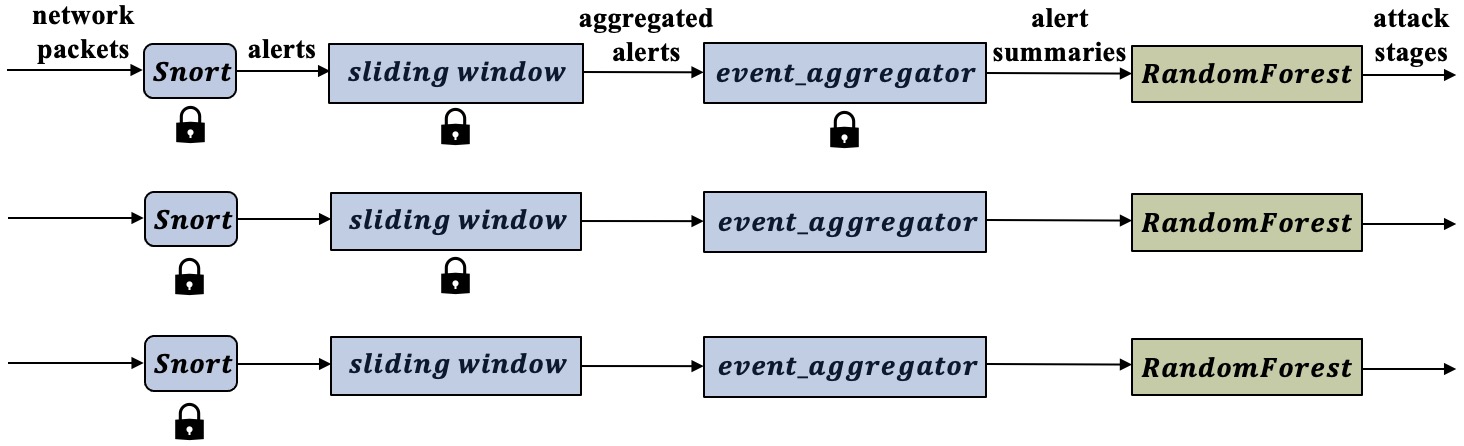}
    }
    \caption{Scenario 3: attack predictors for the MSCAD attack.}
    \label{fig:example3}
  \end{figure}

Figure \ref{fig:example3} shows the components used and how they are connected. In the top chain, the Snort IDS component, the sliding window component, and the event aggregator component (based on an LSTM encoder) are kept fixed during training, while the parameters of the random forest predictor are tuned. In the middle chain, the Snort and the sliding window components remain fixed, whereas the parameters of the event aggregator and the random forest predictor are tuned. Last, in the bottom chain, only the Snort IDS component is pretrained. We train all three chains and evaluate their performance in function of the size of the training data.

\begin{figure}[h]
  \centering
    \scalebox{0.28}{    \includegraphics{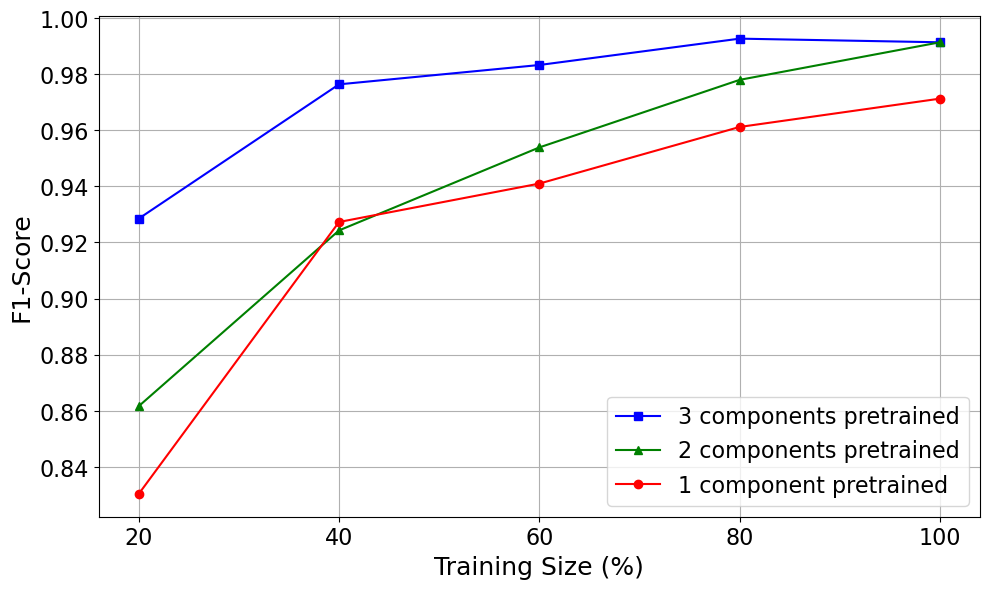}}
    \caption{Prediction accuracy for the MSCAD attack predictors for different sizes of training data. When little data is available, the predictor with more pretrained components performs better.}
    \label{fig:result3}
  \end{figure}

Figure \ref{fig:result3} shows that prediction accuracy improves as more training data becomes available. Furthermore, using a larger number of fixed (i.e., pretrained) components leads to better performance when training data are limited.

\subsection{Scenario 4: controlling training overhead vs. accuracy}
\label{subsec:example4}
In this scenario, we build modular predictors for the CICIDS attack and demonstrate the trade-off between computing resources used for training and prediction accuracy. By controlling the process of candidate predictor training and selection, we regulate the trade-off between composing a more accurate predictor and saving computing resources.

\begin{figure}[h]
  \centering
    \scalebox{0.165}{      \includegraphics{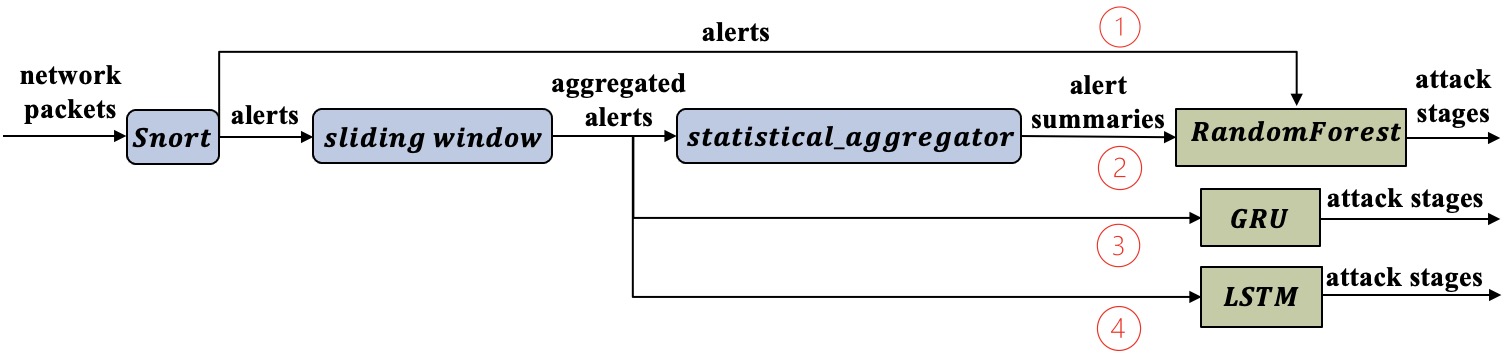}
    }
    \caption{Scenario 4: attack predictors for the CICIDS attack.}
    \label{fig:example4}
\end{figure}

In Figure \ref{fig:example4}, we identify four distinct chains of components, each of which forms an attack stage predictor for the CICIDS attack. The first component of all these predictors contains the Snort IDS component, which produces a sequence of alert events. The first predictor (called Predictor 1) has two components and uses the Snort alerts as input to a random forest classifier, which outputs the attack stage predictions. The other three predictors use a sliding window component, which aggregates the most recent 20 alerts. The second predictor (Predictor 2) feeds the output of the sliding window component to a statistical aggregator and later to a random forest classifier. The third predictor (Predictor 3) feeds the output to a Gated Recurrent Unit (GRU) classifier, and the fourth predictor (Predictor 4) feeds the output to an LSTM classifier.

The goal of Scenario 4 is to build an effective modular predictor from a set of components where we control the training overhead. It is feasible in this case to evaluate all possible predictors and select the best-performing for operation. As mentioned in Section \ref{sec:framework}, this method of evaluation and selection becomes infeasible when the number of components grows significantly. To mitigate this issue, we propose an algorithm (Algorithm 1) to gradually select an effective predictor from a set of candidate predictors during the training process. The algorithm includes a control parameter $k$, for which a smaller value leads to improvement of performance but increases
the training overhead.

\begin{algorithm}[h]
  \label{gradual_algorithm}
  \caption{Selecting a predictor from candidates during training}
  \KwIn{Training dataset $D$, set of candidate predictors $P$, number of folds $k$, performance metric $m$}  
  \KwOut{Final selected predictor $p^*$}
  \While{$|P| > 1$}
  {
    Train each predictor $p \in P$ using $\frac{1}{k}$ from $D$\;
    Rank all predictors in $P$ based on performance metric $m$\;
    Remove the worst-performing predictor from $P$\;
  } 
$p^* \gets$ the remaining predictor in $P$\;
Return $p^*$\;
\end{algorithm}  

Applying Algorithm \ref{gradual_algorithm} to Scenario 4 and setting $k=3$ produces Predictor 2 as the most effective one. This predictor has an accuracy of 0.900 and an F1 score of 0.848. We measure the computational overhead of training as 92 seconds of training time.

For comparison, we train all predictors independently using the full training dataset. Table \ref{tab:example4} shows the prediction accuracy of each predictor. The combined training time of all predictors is 542 seconds. 

We find that both methods select the same attack stage predictor (Predictor 2) with similar performance. However, gradual selection incurs a much lower training overhead (92 vs 542 seconds). While gradual selection is not necessary in this scenario due to the limited number of candidate predictors, it demonstrates how the combinatorial explosion of the number of candidate predictors can be managed as the number of available components increases. We chose Algorithm 1 for illustration purposes, and we are aware that it can be improved. Also, generic algorithms for discrete optimization can be considered in its place.

\begin{table}[h]
\centering
\caption{Prediction accuracy of the four CICIDS attack predictors in Figure \ref{fig:example4} using the full training data.}
\label{tab:example4}
\begin{tabular}{l c r}
\hline
\textbf{Predictor} & \textbf{Accuracy} & \textbf{F1-score}
\\ \hline
Predictor 1  & 0.532    & 0.430   
\\ 
\textbf{Predictor 2}  & \textbf{0.937}    & \textbf{0.897}  
\\
Predictor 3 & 0.907    & 0.845  
\\ 
Predictor 4 & 0.906    & 0.864   
\\ \hline
\end{tabular}%
\end{table}

\section{Related work}
\label{sec:related_work}

We did not find prior work that specifically covers building attack predictors from components. However, several efforts have studied online intrusion prediction based on event streams, the extension of IDSs for automated attack prediction, and trade-offs in intrusion detection. 

\subsection{Online intrusion prediction from processing event streams}

Recent research activities (e.g., \cite{ fu2021realtime,vishwakarma2022dids}) share our objective of predicting attacks in real-time. These works typically focus on processing packet-level events. In \cite{vishwakarma2022dids}, the authors propose a deep neural network-based intrusion detection system for real-time identification of malicious packets. 

When predicting from event streams, window-based event aggregation is a common method. In fact, we have designed such components for event aggregation that we used in several examples in this paper. In \cite{bozdal2021winds}, the authors propose a wavelet-based method for detecting anomalies in controller area network traffic. Consecutive message counts are logged over time windows, and wavelet coefficients are extracted for anomaly detection. In \cite{fu2021realtime}, the authors propose a method for detecting malicious traffic using frequency-domain analysis. Per-packet features are encoded as numerical vectors, segmented into windows, and transformed via discrete Fourier transform to extract temporal features.

Similar to Scenario 1 in this paper, researchers have studied the concurrent processing of multiple event streams, although in the broader context of anomaly detection rather than for attack prediction. Integrating several event streams provides a comprehensive view of the system behavior, often modeled as a multivariate time series \cite{deng2021graph,chen2021learning,zhang2021unsupervised}. In \cite{deng2021graph}, the authors combine deep learning with  GNNs to detect anomalies in multivariate time series. The data are segmented into fixed-length intervals, each representing a static graph where nodes correspond to time series, and edges capture their relationships. A graph attention network models spatial–temporal dependencies, and anomalies are detected when the predicted graphs deviate from the actual graphs. In \cite{zhang2021unsupervised}, the authors propose an unsupervised anomaly detection framework that jointly models spatial and temporal dependencies in multi-sensor time series data. A deep convolutional autoencoder captures spatial relationships among sensors, while an autoregressive model and bidirectional LSTM capture temporal dynamics.

Although the above studies address a problem similar to this paper, they discuss only monolithic predictors. The novelty of these works lies in the careful design of aspects like feature extraction or event aggregation, in order to meet specific requirements. In contrast, our framework isolates such design aspects in form of controllable and reusable components, in order to build predictors for various scenarios and attacks. 

\subsection{Intrusion prediction based on IDSs}
Traditional IDSs like Snort generate low-level alerts indicating potential threats. Similar to our approach, some studies use these alerts as input to train prediction models. In \cite{zhou2021detecting}, the authors propose a sequence-to-sequence model for multi-stage attack detection, where an LSTM encoder transforms alert sequences into latent feature vectors, and an LSTM decoder predicts corresponding attack stages. \cite{wang2024transformer}  introduces a multi-stage attack detection framework. It first employs a similarity-based alert aggregation algorithm to group related IDS alerts and eliminate redundancy. Then, a Transformer-based model is trained on the aggregated alert sequence to capture both local and global dependencies.

A large area of research is dedicated to correlating alerts from IDSs to reduce the rate of alerts that are forwarded to human analysts \cite{lyu2024agcm,mao2021mif,du2024mad,xiao2025grain}. The authors of \cite{lyu2024agcm} and \cite{mao2021mif} propose graph-based approaches that treat IDS alerts as indicators of abnormal network flows. Alerts are modeled as graphs, with nodes representing entities (e.g., IPs or hosts) and edges capturing suspicious relationships, enabling the discovery of behavioral patterns and multi-stage attack sequences. In \cite{xiao2025grain}, the authors present a framework that combines GNNs and reinforcement learning for causality-driven reconstruction of multi-stage attack scenarios. The approach builds alert causal graphs that model the relationships among alerts, uses a GNN to estimate causal effects, and employs a reinforcement learning agent to iteratively refine the graph by removing spurious links and retaining genuine causal paths.

In this paper, we use IDSs as event aggregator components that connect to other components to form a predictor. Our approach goes beyond event reduction, since we predict attack stages. Also, our framework is not limited to IDS alerts as input; it can analyze event streams from various other sources.

\subsection{Trade-offs in intrusion detection}
Operational trade-offs are important aspects to consider in online intrusion detection. A key trade-off lies between achieving accurate predictions and ensuring timely responses. For example, the work in \cite{deng2024robust} addresses website fingerprinting attacks. The proposed method computes correlations between unlabeled traffic and monitored websites in an embedding space, delaying classification when correlations are low and additional data must be gathered for more accurate identification. Another important trade-off concerns saving resources versus increasing detection effectiveness. In \cite{kim2025trade}, the authors propose a traffic sampling framework for intrusion detection in hybrid vehicular networks. By dynamically adjusting the sampling rate based on network conditions, the approach balances the need to capture sufficient traffic data with constraints on bandwidth and processing power, thus achieving effective intrusion detection with small computational overhead.

Additional literature discusses intrusion detection systems that control trade-offs between metrics such as detection accuracy, computational complexity, dimensionality of the input data, and false alarm rate \cite{sharma2024multi}. A line of research addresses such trade-offs through multi-objective optimization, proposing to balance competing goals. Some studies focus on selecting an optimal feature set under constraints on feature set size and prediction accuracy \cite{kareem2022effective,roopak2020multi,mahboob2020anomaly,dey2023hybrid,asgharzadeh2023anomaly}, while others strive for high prediction accuracy with fast convergence during training \cite{wei2020multi,ye2019research,gangula2022network}.

Most of these studies consider a single trade-off and focus on finding an optimal solution under specific constraints. Our framework enables flexible control of various performance metrics, therefore allowing us to regulate a specific trade-off required by a scenario.

\section{Conclusions and future work}
\label{sec:discussion}

We argued in this paper that building monolithic attack stage predictors that cover all attacks in the MITRE framework is not realistic and that a new approach is therefore needed. We proposed a modular framework where predictors are composed from reusable components, and we gave many examples of predictors that have been assembled in this fashion. We found a chain of 2-4 components to be sufficient to build an effective predictor for the examples we studied. We showed that predictor components can be of different internal complexity and can be kept fixed or tunable during training. Moreover, the event streams consumed by each component can be flexibly selected. We demonstrated the controllability of the built predictors across key performance metrics, such as prediction accuracy, timeliness, computational overhead, and sample efficiency, allowing them to be adapted to the diverse requirements of real-world applications.

Regarding future work, much remains to be done to realize our vision of extending and refining the presented framework so that effective attack stage predictors can be efficiently assembled and trained for a given set of attacks through a low-overhead automated process. This work includes identifying a complete (minimal) set of elementary components from which attack stage predictors of interest can be assembled, and it includes making progress on identifying, with low overhead, effective predictors from a large set of candidates, which we discussed in Section \ref{subsec:example4} for a small example. Further topics that warrant investigation in this context include generalizing the framework to accommodate diverse background traffic and infrastructure configurations, providing actionable indications in case of an unknown or zero-day attack, and predicting the progress of an attack described by an attack graph. 

\section{Acknowledgements}
\label{sec:ack}
This work has been supported by the WASP NEST program through project $\mathrm{AIR}^{2}$. The authors thank the $\mathrm{AIR}^{2}$ team members as well as KTH researchers Kim Hammar and Duc Huy Le for their constructive comments.

\bibliographystyle{IEEEtran}
\bibliography{noms2026}

\end{document}